\documentclass[10pt,journal,compsoc]{IEEEtran}

\ifCLASSOPTIONcompsoc
  \usepackage[nocompress]{cite}
\else
  \usepackage{cite}
\fi

\usepackage{amsthm}
\usepackage{amssymb}
\usepackage{graphicx}

\ifCLASSINFOpdf
\else
\fi

\usepackage{amsmath}
\usepackage{algorithmic}

\begin{document}
\title{Computing the Shattering Coefficient of Supervised Learning Algorithms}

\author{Rodrigo Fernandes de Mello, 
        Moacir Antonelli Ponti,
        and~Carlos Henrique Grossi Ferreira
        \IEEEcompsocitemizethanks{\IEEEcompsocthanksitem R. F. de Mello, V. H. C. de Oliveira and M. A. Ponti are with the Department of Computer Science at the Intitute of Mathematics and Computer Science, University of São Paulo.\protect\\
        E-mail: \{mello,vhcandido,moacir\}@icmc.usp.br
        \IEEEcompsocthanksitem C. H. G. Ferreira is with the Department of Mathematics at the Intitute of Mathematics and Computer Science, University of São Paulo.\protect\\
        E-mail: grossi@icmc.usp.br}

}


\IEEEtitleabstractindextext{%
\begin{abstract}
The Statistical Learning Theory (SLT) provides the theoretical guarantees for supervised machine learning based on the Empirical Risk Minimization Principle (ERMP). Such principle defines an upper bound to ensure the uniform convergence of the empirical risk $R_\text{emp}(f)$, i.e., the error measured on a given data sample, to the expected value of risk $R(f)$ (a.k.a. actual risk), which depends on the Joint Probability Distribution $P(X \times Y)$ mapping input examples $x \in X$ to class labels $y \in Y$. The uniform convergence is only ensured when the Shattering coefficient $\mathcal{N}(F,2n)$ has a polynomial growing behavior. This paper proves the Shattering coefficient for any Hilbert space $\mathcal{H}$ containing the input space $X$ and discusses its effects in terms of learning guarantees for supervised machine algorithms.
\end{abstract}

\begin{IEEEkeywords}
Shattering coefficient, Growth function, Supervised machine learning, Theoretical learning guarantees.
\end{IEEEkeywords}}

\maketitle

\IEEEdisplaynontitleabstractindextext
\IEEEpeerreviewmaketitle

\section{Introduction}\label{sec:introduction}

The Statistical Learning Theory (SLT) provides learning guarantees to supervised machine algorithms through the Empirical Risk Minimization Principle (ERMP), defined in Equation~\ref{eq:ERMP} in which $n$ is the sample size, $\epsilon$ sets a divergence factor between the empirical risk $R_\text{emp}(f)$ and its expected value $R(f)$, and $\mathcal{N}(F,2n)$ corresponds to the Shattering coefficient (a.k.a. Growth function)~\cite{Vapnik2013nature}. This coefficient defines the number of distinct classifiers $f \in F$ a supervised learning algorithm is capable of inducing, i.e., the cardinality of the algorithm bias $F$.

\begin{align}\label{eq:ERMP}
    P(\sup_{f \in F} | R_\text{emp}(f) - R(f) | > \epsilon) \leq 2 \mathcal{N}(F,2n) \exp{(-n \epsilon^2 / 4)}
\end{align}

According to the ERMP, the supervised learning only occurs when the upper bound of Equation~\ref{eq:ERMP} converges to zero as $n \rightarrow \infty$. This is only possible if and only if $\mathcal{N}(F,2n)$ is asymptotically dominated by the negative exponential term as follows:

\begin{align*}
    P(\sup_{f \in F} | R_\text{emp}(f) - R(f) | > \epsilon) & \leq 2 \mathcal{N}(F,2n) \exp{(-n \epsilon^2 / 4)}\\
    & \leq 2 \exp{(\log{\mathcal{N}(F,2n)} -n \epsilon^2/4)}
\end{align*}
so that $\log{\mathcal{N}(F,2n)} < n \epsilon^2/4$ as the sample size increases.

Two cases are commonly illustrated in the literature~\cite{Scholkopf2002learning,vonLuxburg}. The first comprehends the scenario in which $\mathcal{N}(F,2n) = n^k$, i.e., the Shattering coefficient is any $k$th-order polynomial (given $k$ is constant), so that:
\begin{align*}
    2 \exp{(\log{\mathcal{N}(F,2n)} -n \epsilon^2/4)}\\
    2 \exp{(\log{n^k} -n \epsilon^2/4)}\\
    2 \exp{(k \log{n} -n \epsilon^2/4)},
\end{align*}
from which we conclude the linear term $-n \epsilon^2/4$ will asymptotically dominate $k \log{n}$ as $n \rightarrow \infty$. The second scenario considers the exponential function $\mathcal{N}(F,2n) = 2^n$:
\begin{align*}
    2 \exp{(\log{\mathcal{N}(F,2n)} -n \epsilon^2/4)}\\
    2 \exp{(\log{2^n} -n \epsilon^2/4)}\\
    2 \exp{(n \log{2} -n \epsilon^2/4)},
\end{align*}
from which no learning guarantee is provided, given $\log{2} > \epsilon^2/4$ once we always set $\epsilon < 1$ to measure the divergence between the empirical and the expected risks.

From this perspective, we conclude the Shattering coefficient $\mathcal{N}(F,2n)$ is essential to prove learning guarantees to supervised machine algorithms. In addition, by having such growth function, we can also find out the minimal sample size to obtain $\epsilon$ as divergence factor (using Equation~\ref{eq:ERMP}). In that sense, this paper presents a proof for the Shattering coefficient for generalized samples contained in any $h$-dimensional Hilbert space, while being classified using a single $(h-1)$-dimensional hyperplane. Next, we formulate this coefficient for any number of $p$ hyperplanes classifying Hilbert spaces.

This paper is organized as follows: the proof of the Shattering coefficient is presented in Section~\ref{sec:computing_shattering}; then, discussions about its impacts are detailed in Section~\ref{sec:discussion}. Finally, conclusions are drawn in Section~\ref{sec:conclusions}.

\section{Shattering coefficient}\label{sec:computing_shattering}

The Shattering coefficient $\mathcal{N}(F,2n)$ of any $h$-dimensional Hilbert space $\mathcal{H}$ being classified with a single $(h-1)$-dimensional hyperplane is:
\begin{align}
    \mathcal{N}(F,2n) = 2 \sum_{i=0}^{h}\binom{n-1}{i},
\end{align}
for a generalized data organization with sample size equals $n$.

In fact, Vapnik and Chervonenkis~\cite{AlexeyChervonenkis} present a slightly different formulation which disconsiders every hyperplane divides a given space into two half spaces, what makes necessary the product of such sum of binomials by $2$. In addition, the numerical results are more precise if we consider binomials in form $\binom{n-1}{i}$ instead of $\binom{n}{i}$ as those authors did.

\begin{figure}
    \centering
    \includegraphics[width=0.7\linewidth]{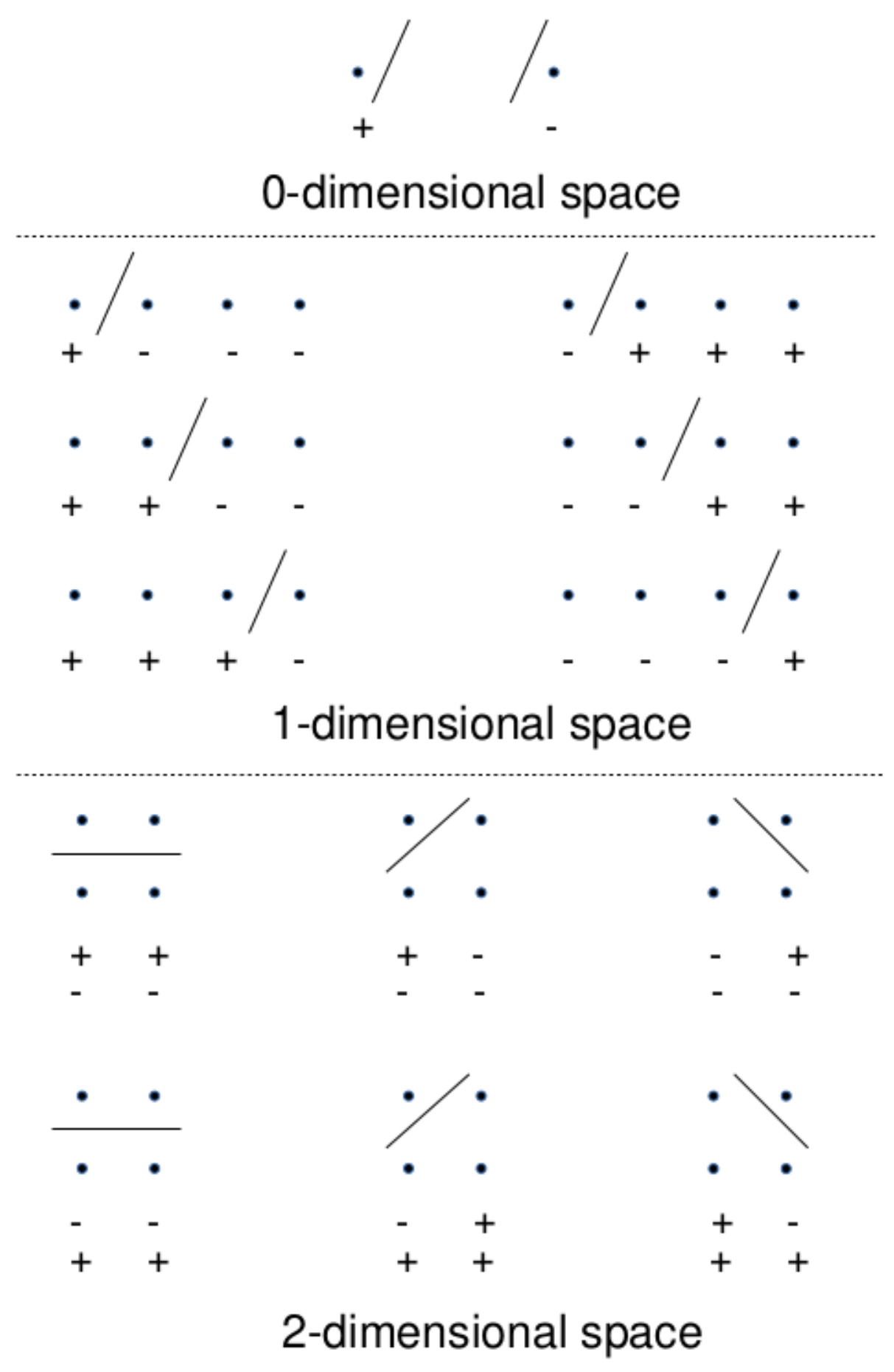}
    \caption{Illustrating the proof on Shattering coefficient.}\label{fig:shattering_illustration}
\end{figure}

\begin{proof}
Let a sample with $2^h$ instances in general organization in a $h$-dimensional Hilbert space $\mathcal{H}$ which must be classified using a single $(h-1)$-dimensional hyperplane. In order to proceed, consider $h=2$ and then that a total of $2^h=4$ instances are firstly projected into a $0$-dimensional Hilbert space, forming a single point (see Figure~\ref{fig:shattering_illustration}). In that scenario, either the point could be classified as laying on one side of the hyperplane or on the other, composing a total of $2$ possibilities (either positive or negative, for example), thus:
\begin{align*}
    2 \binom{4-1}{0} = 2.
\end{align*}

Then, consider the same $2^h=4$ instances are now projected into an $1$-dimensional Hilbert space.
In such data organization, we have six possible classifications, or $2 \binom{4-1}{1} = 6$ in addition to the previous space dimension (see Figure~\ref{fig:shattering_illustration}). Therefore, we have a total of $8$ possible classifications while considering a composition of both spaces, in form:
\begin{align*}
        2 \left( \sum_{i=0}^{1} \binom{4-1}{i} \right) = 8.
\end{align*}
Observe that every time we project the points to a greater dimension, we reorganize them into a generalized form, to next analyze all different classifications obtained when compared to previous spaces. Next, we project points into a $2$-dimensional Hilbert space which can be classified into six other forms ($2 \binom{4-1}{2} = 6$) that remain different from the previous projections, so that:
\begin{align*}
        2 \left( \sum_{i=0}^{2} \binom{4-1}{i} \right) = 14.
\end{align*}
This remains valid for any space dimensionality. In case of any other sample size, this is, if we add a single instance into this current sample with $2^h$, it is obvious that the number of possible classifications can only be equal or greater than the current number of instances, however smaller than the next sample size with $2^h \times 2^{\beta}$ for $\beta \in \mathbb{Z}_+$. From this, we conclude the proof for any $2^h$ sample size as well as the Shattering coefficient is a monotonically increasing function, what is enough to study this coefficient of any hyperplane-based supervised learning algorithm.
\end{proof}

\section{Discussion}\label{sec:discussion}

From our proof of Shattering coefficient, we can conclude that:
\begin{align*}
    \mathcal{N}(F,2n) = 2 \sum_{i=0}^{h}\binom{n-1}{i} = 2^n - 2\sum_{i=h+1}^{n}\binom{n-1}{i},
\end{align*}
therefore, the definition of a $h$-dimensional Hilbert space implies in a reduction of the exponential space of admissible functions $2^n$ of $2\sum_{i=h+1}^{n}\binom{n-1}{i}$. In such a manner, in addition to characterize the cardinality of the algorithm bias, we can also understand its complement to the space containing all possible classifiers for a sample size with $n$ examples.

From this conclusion, we also notice that whenever $h < n$, there is a reduction in the space containing all admissible functions $2^n$, so that such reduction allows to obtain:
\begin{align*}
    &2 \exp{(\log{\mathcal{N}(F,2n)} -n \epsilon^2/4)}=\\
    &2 \exp{\left(\log{\left( 2^n - 2\sum_{i=h+1}^{n}\binom{n-1}{i} \right)} -n \epsilon^2/4\right)},
\end{align*}
as consequence, learning is ensured if and only if (both terms are always positive):
\begin{align*}
    \log{\left( 2^n - 2\sum_{i=h+1}^{n}\binom{n-1}{i} \right)} < n \epsilon^2/4.
\end{align*}

Therefore, when $h \geq n$, we cannot define a more restrictive bias, as consequence the Shattering coefficient is:
\begin{align*}
    \mathcal{N}(F,2n) = 2 \sum_{i=0}^{h}\binom{n-1}{i} = 2^n
\end{align*}
and learning cannot be ensured according the ERMP~\cite{Vapnik2013nature}.

For multiple indexed hyperplanes classifying a given generalized input space, the Shattering coefficient is:
\begin{align*}
    \mathcal{N}(F,2n) = 2 \sum_{i=0}^{h}\binom{n-1}{i}^p,
\end{align*}
due to the direct combination of different hyperplanes, in which $p$ is the number of $(h-1)$-dimensional hyperplanes used to classify the $h$-dimensional Hilbert space $\mathcal{H}$.

As another consequence, we suggest the computation of the Shattering coefficient for supervised learning algorithms in order to prove their uniform convergences and their minimal training set sizes. For instance, suppose the training of an artificial neural network with $p$ neurons, which must produce a classifier $f$ whose empirical risk $R_\text{emp}(f)$ (this may be seen as the risk computed on a test sample) diverges from the actual risk $R(f)$, seen as the risk for unseen data, at most by $5\%$, so that $\epsilon=0.05$ and:
\begin{align*}
&P(\sup_{f \in F} | R_\text{emp}(f) - R(f) | > \epsilon) \leq 2 \mathcal{N}(F,2n) \exp{(-n \epsilon^2 / 4)}\\
&P(\sup_{f \in F} | R_\text{emp}(f) - R(f) | > 0.05) \leq\\
&\;\;\;\;\;\;\;\;\;\;2 \left( 2 \sum_{i=0}^{h}\binom{n-1}{i}^p \right) \exp{(-n \; 0.05^2 / 4)},
\end{align*}
then we may compute the probability in form:
\begin{align*}
    \delta &= 2 \left( 2 \sum_{i=0}^{h}\binom{n-1}{i}^p \right) \exp{(-n 0.05^2 / 4)}.
\end{align*}

Thus, by defining some $\delta$ one wishes to ensure, for example, $\delta=0.01$, one will have a probability of divergence between the empirical and actual risks less than or equal to $0.01$ from which the minimal training set size $n$ can be found:
\begin{align*}
    0.01 &= 2 \left( 2 \sum_{i=0}^{h}\binom{n-1}{i}^p \right) \exp{(-n 0.05^2 / 4)},
\end{align*}
and let $p=16$ so that $16$ hyperplanes are used, and $h=3$, then:
\begin{align*}
    0.01 &= 2 \left( 2 \sum_{i=0}^{3}\binom{n-1}{i}^{16} \right) \exp{(-n 0.05^2 / 4)},\\
    n &\approx 1.02678 \times 10^6.
\end{align*}
which is the training sample size required to ensure such learning guarantee, i.e., in $99\%$ of cases, the empirical risk $R_\text{emp}(f)$ will be a good estimator for the actual risk $R(f)$ given a maximum divergence of $\epsilon=0.05$.

Another interesting conclusion comes from that fact that:
%
%
\begin{align*}
    \left( \frac{m}{k} \right)^k \leq \binom{m}{k} \leq \left( \frac{e m}{k} \right)^k,
\end{align*}
what is valid for $n \ge k > 0$ (proof in Appendix~\ref{app}), then for a single hyperplane:
\begin{align*}
    2 \sum_{i=0}^{h} \binom{n-1}{i} &\leq 2 \binom{n-1}{0} + 2 \sum_{i=1}^{h} \left( \frac{e (n-1)}{i} \right)^i\\
    2 \sum_{i=0}^{h} \binom{n-1}{i} &\leq 2 + 2 \sum_{i=1}^{h} \left( \frac{e (n-1)}{i} \right)^i\\
                                    &\leq 2 + 2 \sum_{i=1}^{h} ( e (n-1) )^i\\
                                    &=\frac{2 e (n - 1) (e^h (n - 1)^h - 1)}{e (n - 1) - 1} + 2,
\end{align*}
and for $p$ hyperplanes:
\begin{align*}
    \sum_{i=0}^{h} \binom{n-1}{i}^p \leq \left( \frac{2 e (n - 1) (e^h (n - 1)^h - 1)}{e (n - 1) - 1} + 2 \right)^{p},
\end{align*}
consequently, we have:
\begin{align*}
    \delta &= 2 \left( 2 \sum_{i=0}^{h} \binom{n-1}{i}^p \right) \exp{(-n \; \epsilon^2 / 4)}\\
           &\leq 2 \exp{\left(\log{\left( \frac{2 e (n - 1) (e^h (n - 1)^h - 1)}{e (n - 1) - 1} + 2 \right)^{p}} -n \; \epsilon^2 / 4 \right)},
\end{align*}
then, from the identity:
\begin{align}\label{eq:log_identity}
    \log_{b}(a+c)=\log _{b}a+\log _{b}\left(1+{\frac {c}{a}}\right),
\end{align}
we have:
\begin{align*}
    \delta &\leq 2 \exp{\left( p \log{\left( \frac{2 e (n - 1) (e^h (n - 1)^h - 1)}{e (n - 1) - 1} + 2 \right)} -n \; \epsilon^2 / 4 \right)}\\
           &=2 \exp{\left(p \log{\left( \frac{2 e (n - 1) (e^h (n - 1)^h - 1)}{e (n - 1) - 1} \right)}\right)}\\
           &\;\;\; \exp{\left(p \log{\left(1 + \frac{2}{\frac{2 e (n - 1) (e^h (n - 1)^h - 1)}{e (n - 1) - 1}} \right)} \right)}\\
           &\;\;\; \exp{(-n \; \epsilon^2 / 4)},
\end{align*}
and as $n \rightarrow \infty$:
\begin{align*}
    \lim_{n \rightarrow \infty} \frac{2}{\frac{2 e (n - 1) (exp^2(1) (n - 1)^2 - 1}{e (n - 1) - 1}} = 0,
\end{align*}
then, we have:
\begin{align*}
    &\exp{\left(p \log{\left(1 + \frac{2}{\frac{2 e (n - 1) (e^h (n - 1)^h - 1)}{e (n - 1) - 1}} \right)} \right)} =\\
    &\exp{\left(p \log{\left(1 + 0 \right)} \right)} =\\
    &\exp{(\log{(1)})} = 1, \;\; \text{as}\; n \rightarrow \infty,
\end{align*}
then:
\begin{align*}
    \delta &\leq 2 \exp{\left(p \log{\left( \frac{2 e (n - 1) (e^h (n - 1)^h - 1)}{e (n - 1) - 1} \right)} -n \; \epsilon^2 / 4\right)},
\end{align*}

Now let:
\begin{align*}
    \psi &= p \log{\left( \frac{2 e (n - 1) (e^h (n - 1)^h - 1)}{e (n - 1) - 1} \right)} -n \; \epsilon^2 / 4.
\end{align*}
In order to make the exponential function converge to zero, the term $\psi$ must be negative ($\psi < 0$):
\begin{align*}
    p \log{2} &+ p \log{e} + p \log{(n - 1)} + p \log{(e^h (n - 1)^h - 1)}\\
              &-p \log{(e (n - 1) - 1)} - n \epsilon^2/4 < 0,
\end{align*}
from which we obtain the following after also applying the identity in Equation~\ref{eq:log_identity}:
\begin{align*}
    p \log{2} + h p + h p \log{n} - n \epsilon^2/4 < 0,
\end{align*}
as $n \rightarrow \infty$, which is the general-purpose upper bound formulation for any $h$-dimensional Hilbert space being classified using $p$ hyperplanes. 

\begin{figure}
    \centering
    \includegraphics[width=1.0\linewidth]{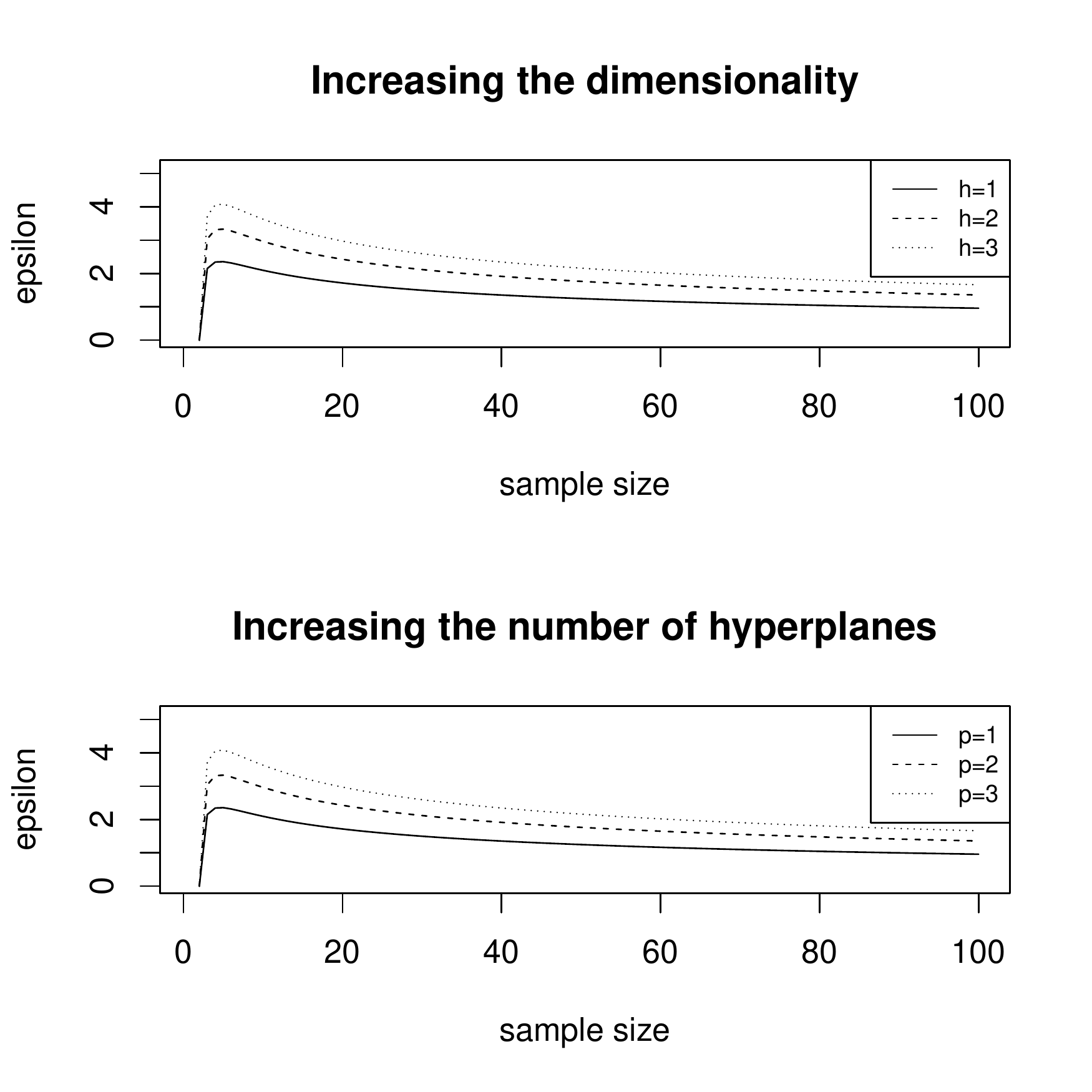}
    \caption{Impacts on the divergence $\epsilon$ as the numbers of hyperplanes $p$ and dimensions $h$ of the Hilbert space grow.}
    \label{fig:dimensions}
\end{figure}

In addition, given $h$ and $p$ are constants (defined according to the target problem), the term $p \log{2} + h p = \gamma$ is also a constant here referred to as $\gamma$, therefore the convergence depends on:
\begin{align*}
     h p \log{n} - n \epsilon^2/4 < -\gamma.
\end{align*}
We here rewrite this inequation as follows:
\begin{align*}
    h p \log{n} = n \epsilon^2/4 - \gamma,
\end{align*}
to study how difficult is to ensure some divergence $\epsilon$ as the number of hyperplanes $p$ and the number of dimensions $h$ grow (observe Figure~\ref{fig:dimensions} assesses the value of $\epsilon$ as the sample size increases):
\begin{align*}
    \epsilon = \frac{2 \sqrt{h p \log{n} + \gamma}}{\sqrt{n}}.
\end{align*}
Curves confirm an upward offset in $\epsilon$ whenever $p$ or $h$ are increased, making evident this divergence becomes more difficult to be ensured when any of those parameters grows. This is due to the obtained spaces contain more admissible functions, i.e., the algorithm bias is less restricted.

\section{Acknowledgments}

We kindly thank Victor Hugo C\^{a}ndido de Oliveira for revising the formulation.

\section{Conclusions}\label{sec:conclusions}

This paper proves the Shattering coefficient for any $h$-dimensional Hilbert space being classified using a $(h-1)$-dimensional hyperplane. As result, we also show how to compute this coefficient when multiple hyperplanes are considered. From that, we also draw conclusions including how researchers and end-users can employ the Empirical Risk Minimization Principle to find the minimal training sample size to ensure learning bounds for general-purpose supervised machine algorithms.

\bibliographystyle{plain}

\appendices
\section{Proof of lower and upper bounds for binomial coefficients}\label{app}

\noindent \textbf{Lower bound:} the following result is known:
\begin{align*}
    \left( \frac{n}{k}\right)^k \leq \binom{n}{k} \leq \left( \frac{e n}{k}\right)^k,
\end{align*}
for $n \geq k > 0$. The proof starts with $k=1$, from which we have:
\begin{align*}
    \left( \frac{n}{k} \right)^k = n \leq n = \binom{n}{k}.
\end{align*}
Then, considering $k > 1$ and let $0 < m < k \leq n$:
\begin{align*}
    k \leq n \Rightarrow \frac{m}{n} \leq \frac{m}{k} \Rightarrow 1-\frac{m}{k} \leq 1-\frac{m}{n}\\
    \;\;\Rightarrow \frac{k-m}{k} \leq \frac{n-m}{n} \Rightarrow \frac{n}{k} \leq \frac{n-m}{k-m},
\end{align*}
so that:
\begin{align*}
    \left( \frac{n}{k} \right)^k = \frac{n}{k} \cdot \ldots \cdot \frac{n}{k} \leq \frac{n}{k} \cdot \frac{n-1}{k-1} \cdot \ldots \cdot \frac{n-k+1}{1}=\binom{n}{k}.
\end{align*}

\noindent \textbf{Upper bound:} the following result is known:
{\small
\begin{align}\label{eq:proof:upper:bound}
    \binom{n}{k} = \frac{n!}{(n-k)! k!} = \frac{n(n-1)\ldots(n-(k-1))}{k!} \leq \frac{n^k}{k!}.
\end{align}}
Then, from the following expansion:
\begin{align*}
    e^{k} = \sum_{i=0}^{\infty} \frac{k^i}{i!},
\end{align*}
if one selects a term $k$:
\begin{align*}
    e^{k} > \frac{k^k}{k!},
\end{align*}
implying:
\begin{align*}
    \frac{1}{k!} < \binom{e}{k}^k.
\end{align*}
Substituting this last term in Equation~\ref{eq:proof:upper:bound}, the bound is proven.

\begin{IEEEbiography}[{\includegraphics[width=1in,height=1.25in,clip,keepaspectratio]{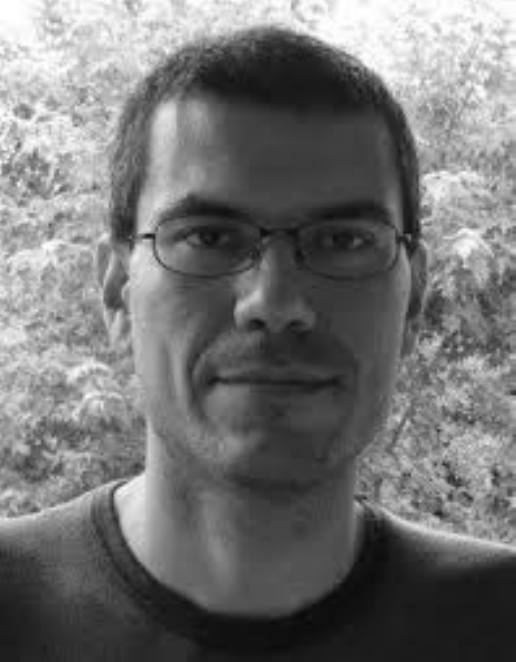}}]{Rodrigo Fernandes de Mello} is Associate Professor with the Department of Computer Science, at the Institute of Mathematics and Computer Sciences, University of S\~{a}o Paulo, S\~{a}o Carlos, Brazil. He obtained his PhD degree from University of S\~{a}o Paulo, S\~{a}o Carlos in 2003. His research interests include the Statistical Learning Theory, Machine Learning, Data Streams, and Applications in Dynamical Systems concepts.
\end{IEEEbiography}

\begin{IEEEbiography}[{\includegraphics[width=1in,height=1.25in,clip,keepaspectratio]{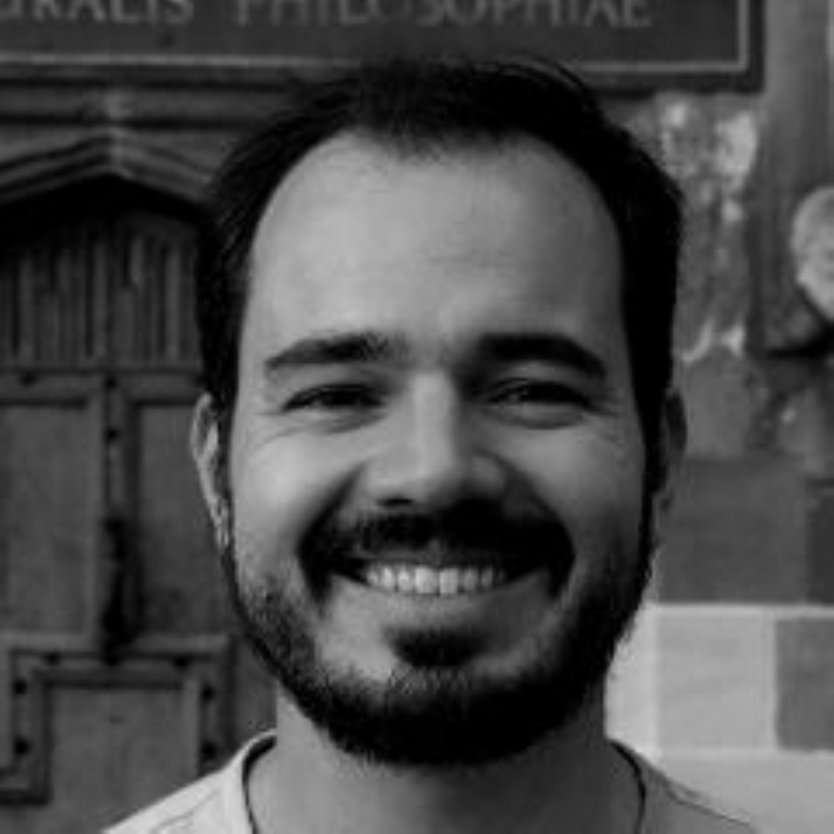}}]{Moacir Antonelli Ponti} is Associate Professor with the Department of Computer Science, at the Institute of Mathematics and Computer Sciences, University of São Paulo, São Carlos, Brazil, and visiting professor at the Centre for Vision, Speech and Signal Processing (CVSSP), University of Surrey in 2016. He obtained his PhD from the Federal University of S\~{a}o Carlos. His research interests include Multiple-classifiers systems, Biomedical signal processing, Image restoration and Visual Pattern Recognition.
\end{IEEEbiography}

\begin{IEEEbiography}[{\includegraphics[width=1in,height=1.25in,clip,keepaspectratio]{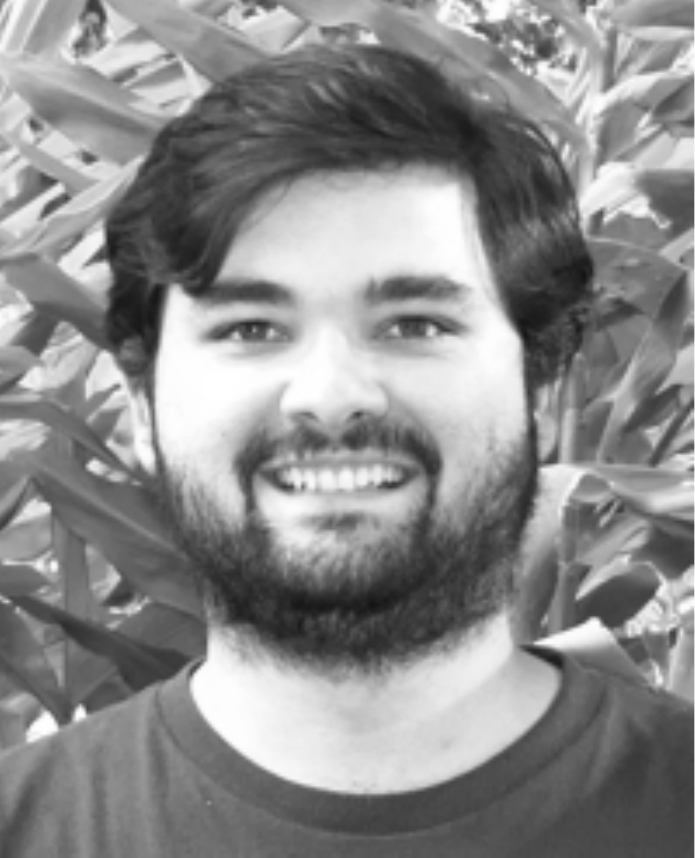}}]{Carlos Henrique Grossi Ferreira} is Associate Professor with the Department of Mathematics, at the Institute of Mathematics and Computer Sciences, University of São Paulo, São Carlos, Brazil. He obtained his PhD from the State University of Campinas, Brazil. His research interests include Riemannian geometry, Hyperbolic and Hyperbolic complex geometries.
\end{IEEEbiography}

\end{document}